\documentclass{article}
\usepackage[english]{babel}
\usepackage[letterpaper,top=2cm,bottom=2cm,left=3cm,right=3cm,marginparwidth=1.75cm]{geometry}

\usepackage{amsmath}
\usepackage{graphicx}
\usepackage[colorlinks=true, allcolors=blue]{hyperref}

\title{Addressing Issues with Working Memory\\ in Video Object Segmentation}
\author{Clayton Bromley, Alexander Moore, Amar Saini, Douglas Poland, Carmen Carrano\\
\small{\{bromley1, moore278, saini5, poland1, carrano2\}@llnl.gov}}

\begin{document}
\maketitle

\begin{figure}[h!]
\centering
\includegraphics[width=0.70\linewidth]{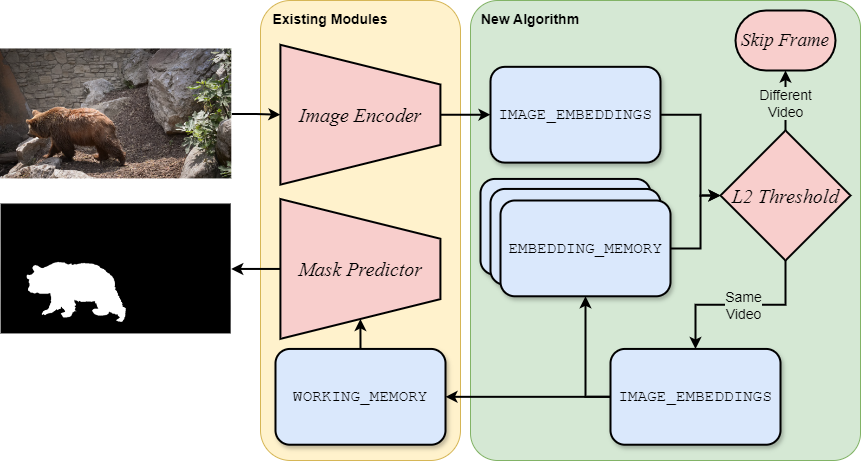}
\caption{\label{fig:method} General Interjection Classifier}
\end{figure}

\begin{abstract}

\noindent Contemporary state-of-the-art video object segmentation (VOS) models compare incoming unannotated images to a history of image-mask relations via affinity or cross-attention to predict object masks. We refer to the internal memory state of the initial image-mask pair and past image-masks as a working memory buffer. While the current state of the art models perform very well on clean video data, their reliance on a working memory of previous frames leaves room for error. Affinity-based algorithms include the inductive bias that there is temporal continuity between consecutive frames. To account for inconsistent camera views of the desired object, working memory models need an algorithmic modification that regulates the memory updates and avoid writing irrelevant frames into working memory. A simple algorithmic change is proposed that can be applied to any existing working memory-based VOS model to improve performance on inconsistent views, such as sudden camera cuts, frame interjections, and extreme context changes. The resulting model performances show significant improvement on video data with these frame interjections over the same model without the algorithmic addition. Our contribution is a simple decision function that determines whether working memory should be updated based on the detection of sudden, extreme changes and the assumption that the object is no longer in frame. By implementing algorithmic changes, such as this, we can increase the real-world applicability of current VOS models.

\end{abstract}

\section{Introduction}

VOS is an ongoing challenge in the world of video processing and understanding. As models continue to improve performance on clean data, steps must be taken to increase robustness towards challenges found in real-world video data that clean benchmarks fail to capture. One such augmentation that can be expected in both streaming and recorded video data is a camera cut: any instance in which the context changes so significantly from one frame to the next that the segmented object is reliably absent. This can occur by way of sudden camera movement, such that the new frame is entirely different from the previous, or the splicing of an entirely different video. VOS datasets and benchmarks are consistently high-quality and continuous video streams, while in the real world, cuts and scene changes introduce an additional layer of complexity to which the inductive biases for current VOS models fail to generalize.\\

\noindent Current and previous state-of-the-art VOS models have significant performance drops when these camera cuts are present in the data. When irrelevant frames are present in video data, the following process occurs:

\begin{enumerate}
    \item The desired object is segmented prior to a camera cut.
    \item A camera cut or discontinuity occurs, and the model suddenly loses track of the object. Object attention disappears and current frame-to-frame smooth motion is interrupted.
    \item The model may latch onto a different object in the post-camera cut frame, resulting in false positive mask predictions. This is because models undergo training on VOS standard data and benchmarks, where objects are always continuous through their spacetime positions.
    \item The subsequent irrelevant frames are written into the model's object memory.
    \item The object memory deteriorates over time as new irrelevant memories populate the buffer.
    \item If the object reappears, the re-identification is ineffective because of the memory deterioration during the irrelevant frames.
\end{enumerate}

\noindent We propose a simple algorithmic addition to improve performance on data with camera cuts. By implementing a binary classifier via L2 distance on image embeddings from frame to frame, it is possible to determine when camera cuts occur. Then, false positive predictions can be prevented, and irrelevant frames can avoid being written to memory. This switch maintains object attention despite long camera cuts and substantially improves re-identification during long intermission periods from relevant objects.

\subsection*{Contributions}

\noindent In this paper we improve former and current state-of-the-art VOS models on multiple benchmarks involving temporal inconsistencies using a model-agnostic approach. This includes the following steps:

\begin{enumerate}
    \item We demonstrate that frame interjections are detectable without fine-tuning image encoders for previous three state-of-the-art VOS models.
    \item We use the image embedding stream and sequence knowledge to compare frame-to-frame embeddings by proposing a new distance function on the image embedding space utilizing sequence for standard error.
    \item We engineer features which lead to highly-discriminatable interjection periods.
    \item We propose a simple classifier to determine interjections and improve regulation of the working memory buffer on the three previous state-of-the-art VOS models.
\end{enumerate}

\section{Related Works}
VOS models generally employ one of two different methods for mask prediction. Until recently, state-of-the-art models were affinity-based, including XMem\cite{XMEM} which first applied the concept of an external working memory buffer to the VOS task. Segment-Anything-Model 2\cite{SAM2} (SAM 2) introduced an architecture which depends only on memory embeddings rather than previous frames, and provides improvement over other architectures. It is still, however, limited by the working memory assumption that frames are continuous and relevant.

\subsection{Affinity-Based VOS}

Some VOS models rely on the affinity, or soft-max similarity, between a frame's mask/image embeddings, and the keys stored in memory from previous frames. For example, XMem\cite{XMEM}, a VOS model released in 2022, implements affinity alongside a three-tiered memory system. Upon receiving an initial mask and frame, a key encoder and a value encoder are used to create an embedding representation of the frame and the image-mask relationship respectively. These embeddings are written to the working memory layer as a function of the affinity to previous frame embeddings. However, with each progressive memory layer, the memory is encoded after a constant amount of time (default 5 frames). As such, the model is unable to account for irrelevant frames. The affinity matrix between the memory and the mask/frame embeddings is then used to generate the new predicted segmentation mask.\\

\noindent Cutie\cite{CUTIE} is an improved affinity-based VOS model that was built from XMem and utilizes the same pixel memory structure. First, Cutie encodes segmented frames into a high-resolution pixel memory and a high-level object memory to be stored for future use. The same working memory buffer structure is used in both XMem and Cutie, and thus they are both susceptible to the same data imperfections. An initial pixel readout is retrieved from the pixel memory using encoded query features and enriched with object-level semantics by augmenting with information from the object memory. The enriched output is finally passed to the decoder for generating the final output mask.\\

\noindent While these affinity-based models consider the frame-to-frame similarity in mask prediction, there are no checkpoints in place to directly account for sudden changes, such as camera cuts. We propose a fix that applies the affinity between the embeddings of the current frame and the previous frame to create a binary "classifier to identify camera cuts and avoid writing irrelevant frame embeddings into memory.

\subsection{Affinity-Free VOS}

Unlike its predecessors, Segment-Anything-Model 2\cite{SAM2} (SAM 2) is not dependant on affinity, and it fundamentally changes the way VOS is explored as a problem. Rather than using frame-to-frame affinity for mask prediction, SAM 2 utilizes attention between the frame embeddings and the memory, to which we refer as "affinity-free modeling". This model can be seen as a generalization of the original segment-anything-model\cite{SAM}, the state-of-the-art image segmentation model, such that images can be explored as one frame videos. Masks, bounding boxes, and points can all be used as inputs at any point through a video, and the prompt encodings are using to decode the mask prediction. Finally, the predicted mask is encoded into memory for the subsequent mask prediction.\\

\noindent Unlike XMem and Cutie, SAM 2 adds an additional head to the mask decoder to predict whether the object is visible in the frame via a multilevel perceptron (MLP). The occlusion head determines the probability that the object is not present in the frame and writes the memory accordingly. This new addition does an excellent job at accounting for objects that are slowly and temporarily obscured. As shown in Table \ref{table:full}, however, sudden and extreme cuts or context shifts still can cause SAM 2 to lose track of the segmented object.

\subsection{Datasets}

Previous datasets have been created to explore the segmentation of partially or fully obscured objects. DAVIS\cite{DAVIS} is a dataset of multi-object videos which have been segmented for ground truth masks. It is a highly popular benchmark on which models can compare performance. While this data contains limited object obscurations, it provides a good baseline for VOS. The DAVIS dataset will be used to generate the data used in this paper.\\

\noindent MOSE\cite{MOSE} is a popular video dataset similar to DAVIS that emphasizes object obscuration. In many samples, obscured objects will become partially or fully obscured for a section of the video to test VOS model robustness. State-of-the-art models, particularly Cutie and SAM 2, have emphasized this dataset when developing their architectures and have great performance on these obscurations. However, MOSE fails to provide fully-annotated long-term scalable obscurations and sudden camera cutting. For this reason, we use DAVIS to create a dataset of videos with the necessary annotated long-term interjections.

\section{Experimental Design}

\begin{figure}[h!]
\centering
\includegraphics[width=1\linewidth]{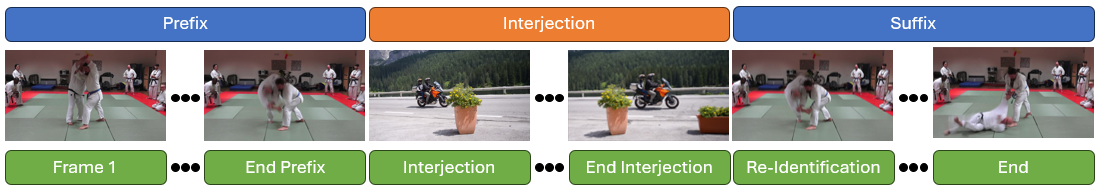}
\caption{\label{fig:interjection_video}Interjection Video Data Structure}
\end{figure}

An interjection dataset is created from DAVIS to benchmark the various models explored in this paper. An object is selected from a video of interest, and a select number of frames from a separate, unrelated video are interjected in the middle of the video of interest, as shown in Figure \ref{fig:interjection_video}. An ideal VOS model is expected to do the following during the three stages of the interjection video:

\begin{enumerate}
    \item \textit{Prefix}: Take the initial mask and write the object into memory
    \item \textit{Interjection}: Identify the absence of the object and make no false positive predictions
    \item \textit{Suffix}: Re-identify the object and continue segmenting at the same accuracy as during the prefix.
\end{enumerate}

\noindent Five datasets are collected and benchmarked over all explored VOS models. First, the clean DAVIS video dataset is benchmarked to ensure that any modifications designed to improve performance on camera cuts does not diminish performance on reliable, clean video. Then, four separate datasets are created using the structure in Figure \ref{fig:interjection_video}. In all cases, the prefix and suffix are both 12 frames to prevent biased results from having additional time to embed the object of interest into memory. The interjection lengths are 4, 16, 128, and 512 frames respectively, allowing the models to be benchmarked on both short-term and long-term obscurations.

\section{Method}

Working-memory-buffer utilizing models have no decision function on updating the working memory. Rather, memory is updated every 5 steps by default. Our algorithmic modification introduces a learnable step in which the model uses image features to determine whether the memory should be updated to move away from strong assumptions made for VOS benchmarks on clean data and toward more challenging real-world data. The general method employed to account for this data is shown in Figure \ref{fig:method}. The result is a simple binary classifier that identifies whether or not each consecutive frame is part of the same video or if a context change has occurred. Each frame is encoded using the encoder built into the model, which is trained around the segmentation task, and the resulting image embeddings are compared to the image embeddings from previous frames. If the features are highly dissimilar in segmentation-embedding space, it means the shape of the objects present in the scene are highly different, and indicator of a context shift\\

\noindent The L2 distance between the current embeddings and the element-wise z-score of the previous frame embeddings is used as the similarity function as shown in Equation 1 where $f_i$ represents the encoded pixel values of the current frame in stream, $f_m$ represents the element-wise mean of the equivalent encodings for the frames in memory, and $\sigma_m$ represents the element-wise variance of the memory encodings from the previous frames.

\begin{equation}
   \text{Regularized Distance} = \text{L2}\left( \frac{f_i - f_{m}}{\sigma_m} \right)
\end{equation}

\noindent The memory frame embeddings come from a stored list of image embeddings that are known to be from the original video. While we know the true placement of interjection frames during training, this placement is unknown during validation. For a context window of size w, the previous w frame image embeddings are used to find the element-wise mean ($f_m$) and element-wise variance ($\sigma_m$) such that $f_i$, $f_m$, and $\sigma_m$ all maintain the same dimensionality. A context window is necessary to avoid overfitting to frames with small variances over long video lengths. For example, without a context window, CCTV video footage located in an inactive area might overfit to no variance between frames, then classify an interjection whenever anything changes due to the large resulting Z-score.\\

\noindent The resulting L2 distance values can be compared to a threshold, frame by frame, to complete the classification without the need for any trained parameters. If the frame is below the threshold, then it can be classified as a part of the same video, and the model's mask predictor can operate as usual. The image embeddings can be written to working memory and stored for future frame embeddings comparisons. If the frame is above the threshold, it is assumed to be an interjection, the current position is not added to memory buffer, and the mask predicts no false positives. The challenge is in finding a threshold function that identifies interjection frames 100\% of the time without any false positive classifications. Several functions are explored to accomplish this.

\subsection{Zeroth-Order L2 Comparison}

The simplest threshold possible is a set value that acts as the classification decision boundary. Any L2 values above the threshold value is an interjection while any below is from the same video. This function, while easy to implement, is unreliable for all context window lengths. In cases where the interjected video is conceptually similar to the original video, the L2 value might be small despite being an interjection. For example, if a video of a animal is interjected with a video of a different animal, the embedding difference may be below the threshold.\\

\noindent Similarly, if a video contains quick movements or has a low frame rate, the embedding differences might be too large, resulting in a false positive classification. Because the lowest possible interjection L2 value is below the highest possible non-interjection L2 value, zeroth-order comparisons cannot be used for accurate classifications.

\subsection{First-Order L2 Comparison}

The first-order threshold function takes into account comparisons between the current L2 distance and the previous window L2 distance. Both the difference between the values and the ratio between the values are explored alongside a zeroth-order comparison to create a piecewise threshold function with significantly higher accuracy.\\

\begin{figure}[h!]
\centering
\includegraphics[width=0.90\linewidth]{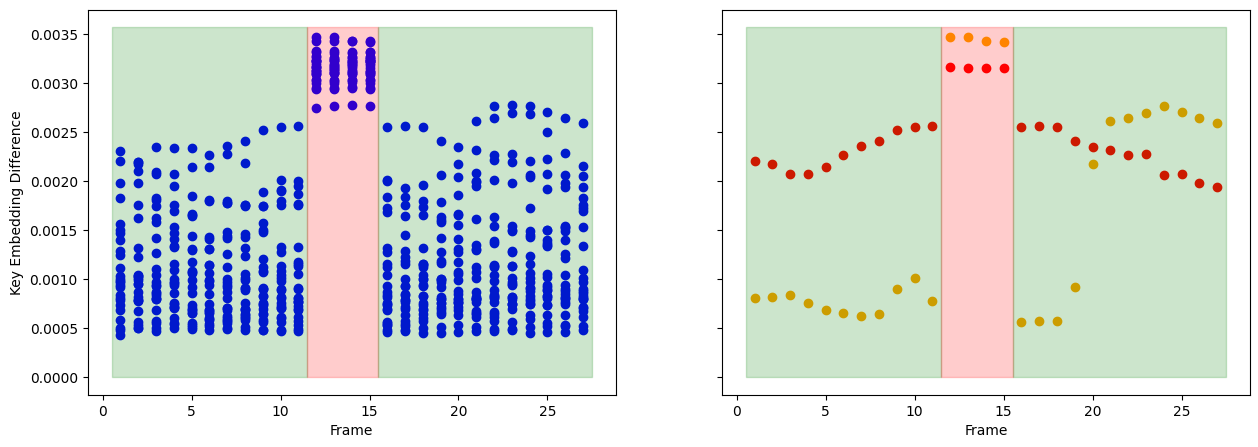}
\caption{\label{fig:first-order} Frame-to-Frame L2 Distances over 25 Samples}
\end{figure}

\noindent The first plot in Figure \ref{fig:first-order} shows the image embedding L2 distances for every frame (w=1) in 25 interjection video samples. The points in the green area represent frames that are from the same video while the points in the red area represent interjected frames. The samples shown here have a 12 frame prefix, 4 frame interjection (from frames 12-15), and a 12 frame suffix.

\noindent While nearly every sample can be correctly classified using these techniques, a few samples in particular cause difficulties. Two of these samples are shown in the second plot of Figure \ref{fig:first-order}. The red sample in the second plot has a relatively small jump from frame 11 to frame 12. In this particular sample, the interjection frames are contextually similar to the prefix frames so functions that rely on L2 difference and ratio may cause the model to not recognize the interjection. The orange sample has the opposite issue. The jump between the prefix, interjection, and suffix are relatively large and therefore easy to identify. Between frames 19 and 20, however, there is another large jump caused by quick camera movements. First-order threshold functions tend to incorrectly identify this jump as an interjection. It is impossible to create a perfect classifier using this technique without including many piecewise components, however this results in overfitting and cannot be generalized.\\

\noindent Another issue with this technique is the number of frames required to make a decision. Each L2 distance requires w+1 frames to compute. Thus, comparing L2 distance values between windows requires a minimum of w+2 frames. If an interjection occurs within the first w+2 frames of a video, it will be impossible for this technique to identify it. Higher-order classifiers will continue to magnify this problem so a different approach is necessary.

\subsection{Maximum Distance Ratio}

The model can rely on the fact that frames in the same video are likely to be far more similar to each other than to an interjected frame. Regardless of the actual L2 distance values between the frames, it is a safe assumption that the distance to the interjected frame is larger than any of the distances between any of the frames within the same video. This can be used to avoid false positive interjection classifications. In cases where the camera quickly pans, the distance to the subsequent frame will be large despite being from the same video. Yet, the distance will still be smaller than an interjection in the same sample.\\

\noindent This method maintains a variable for the maximum window distance seen up to that point. For each new window explored, the maximum distance ratio (MDR) is the window L2 distance divided by the maximum distance seen. Thus, if a frame is furthest from the previous frames, the MDR will be greater than 1 and vice versa. Large MDR values are indicative of interjection frames. Once an interjection is identified, subsequent interjection frames will compute MDR values in comparison to the initial interjection. As a result, the remaining MDR values will be around 1. For this reason, MDR cannot be the only indicator of interjection, but when used in conjuction with the other methods, accuracy can be maximized.\\

\begin{figure}[h!]
\centering
\includegraphics[width=0.50\linewidth]{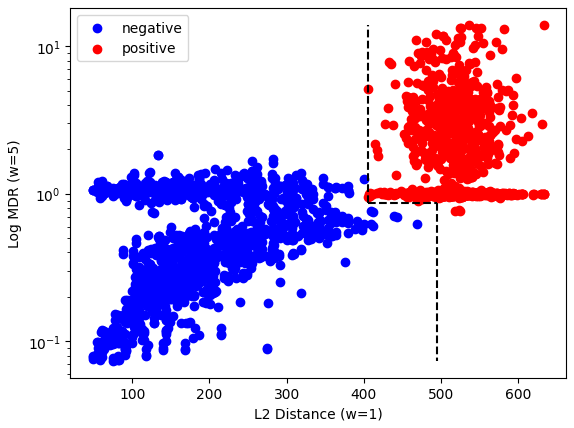}
\caption{\label{fig:MDR} Decision Boundary using MDR and Zeroth-Order L2}
\end{figure}

\noindent Figure \ref{fig:MDR} demonstrates the benefit of using MDR. This plot shows all frame windows (w=5) in a dataset in which the first-order L2 distance ratio are greater than 1.07. There are no interjection frames that have a distance less than 1.07 times the previous window. In this figure, the window MDR is  plotted against the raw L2 distance from the previous frame (w=1). Interjections are shown in red while non-interjections are shown in blue. A relatively simple decision boundary can be formed to achieve a perfect classifier, as shown by the black dotted line. While MDR is imperfect by itself, it can be used alongside the other methods to create highly accurate classifiers.

\section{Implementation}

This injection classification technique was applied to Cutie, XMem, and SAM 2 VOS models. Both long-term distance (w=5) and short-term distance (w=1) are explored. To prevent the loss of frames before a full long-term window, intermediate windows of size [1, 2, ... , w-1] are used on the first w-1 frames instead.

\subsection{Cutie}

\begin{figure}[h!]
\centering
\includegraphics[width=0.50\linewidth]{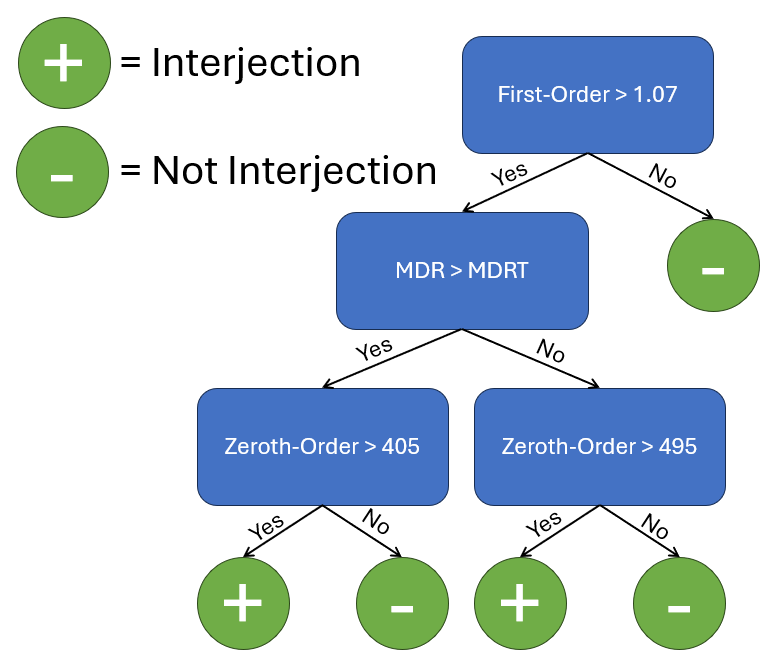}
\caption{\label{fig:cutie_tree} Decision Tree for Cutie Threshold}
\end{figure}

One of the Cutie interjection datasets is represented in Figure \ref{fig:MDR}. The tree used to make the interjection classification is shown in Figure \ref{fig:cutie_tree}. The long-term first-order ratio of the current window divided by the previous window is first used to remove any obvious non-interjection frames. The L2 distance must be decently large compared to the frames known to be in the same video. Then the MDR value is computed using the same long-term context window and compared to a MDR threshold (MDRT) that varies based on the length of an interjection. Finally, the short-term zeroth-order raw L2 distance is used to make the final decision, as shown in the diagram.\\

\noindent As the length of an interjection increases, so does the variation of interjection frames. Some interjection frames are closer to the non-interjected frames than others, and the MDR will reflect this. For long interjections there is an increased likelihood that an interjected frame will have a lower MDR, since it is compared to the most dissimilar interjection frame. Thus, the MDRT must be a function of the length of an interjection. For simplicity, a linear MDRT function is shown in Equation 2 where \textit{l} represents the current length of the interjection. MDRT decreases linearly with interjection length for the first 120 frames of an interjection before minimizing at 0.50. This function returned perfect classification results for the used datasets (up to 512-frame interjections), however a more nuanced decay function may provide better results.

\begin{equation}
   \text{MDRT} = \max(0.86-0.003l, 0.50)
\end{equation}

\subsection{XMem}

\begin{figure}[h!]
\centering
\includegraphics[width=1\linewidth]{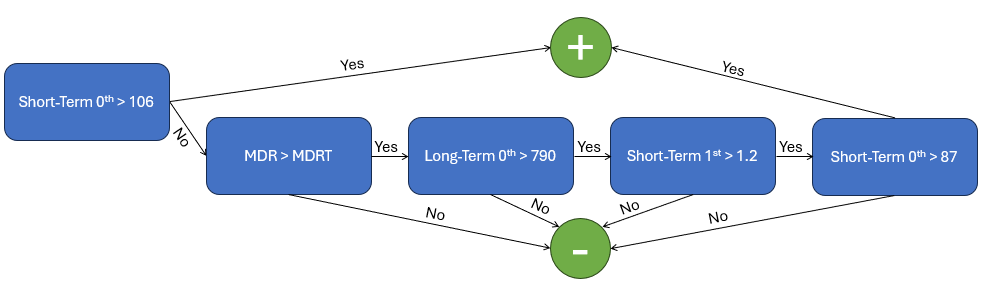}
\caption{\label{fig:xmem_tree} Decision Tree for XMem Threshold}
\end{figure}

Because XMem is an older model than Cutie, the frame encoder is of worse quality. Thus, additional steps are required to achieve high accuracy in interjection classification. Mainly, the zeroth and first order threshold function are explored for both the short term frame window (w=1) and the long term frame window (w=5). The same MDRT function is used, and as with the Cutie decision tree, MDR is computed using the long-term window. The decision tree used is shown in Figure \ref{fig:xmem_tree}.

\subsection{SAM 2}

While the SAM 2 algorithm is affinity free, it still relies on working memory and is thus susceptible to being warped by irrelevant frames. Because of the more complex encoding algorithm used by SAM 2, the same thresholding methods cannot be used by themselves without overfitting to the training data. Alternatively, a variety of functions were used to effectively separate the available interjection and non-interjection frames. Equation 3 shows the best thresholding function identified, where $ST_0$ and $ST_1$ represent the zeroth-order and first-order short-term thresholding functions respectively. Above this value, all points are correctly identified as interjections.

\begin{equation}
    ST_0 * ST_1 > 287
\end{equation}

\noindent Simple functions such as $ST_0 > 170$, $ST_1 > 1$, and MDR $> 0.97$ can also be used to identify interjections with perfect accuracy. The remaining points, however, cannot be perfectly separated using the currently explored methods. Equation 4 can be used to classify the general trend of interjection frames with 97.6\% accuracy, but additional research is required to find a perfect SAM 2 interjection classifier.

\begin{equation}
   ST_1 > e^{-0.15\left(ST_0-170\right)} + 1.03
\end{equation}

\noindent Long-term frame similarity (w=5) is not a useful classifier for SAM 2, as the previous frame is a better indicator of interjection in both zeroth and first order L2 functions.

\section{Results}

\begin{figure}[h!]
\centering
\includegraphics[width=1\linewidth]{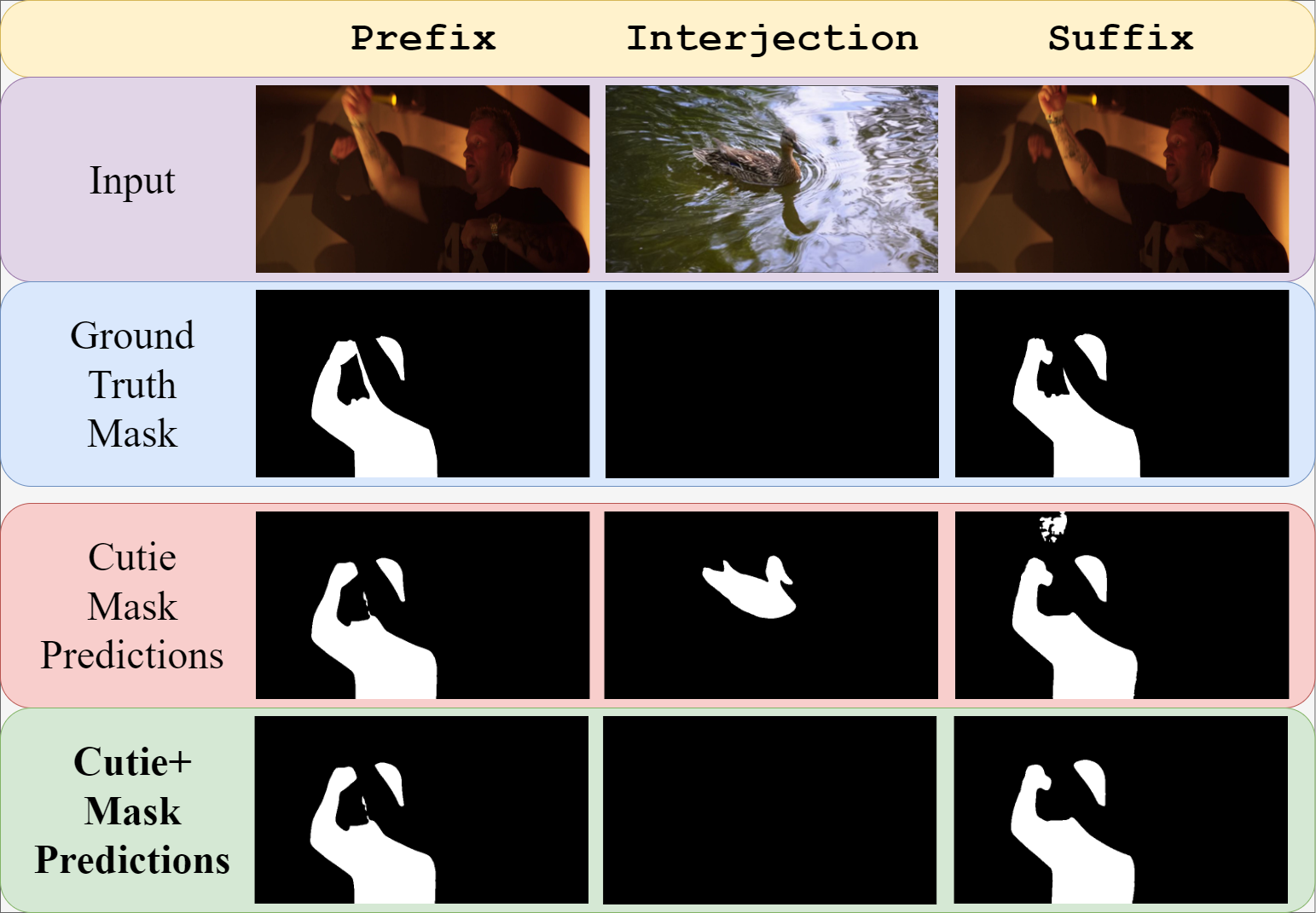}
\caption{\label{fig:cutie pics} 512-Frame Interjection Video Mask Predictions for Cutie and Cutie+}
\end{figure}

\noindent Figure \ref{fig:cutie pics} shows a visualization of the mask predication for a 512-frame interjected video applied to Cutie and Cutie+. Vanilla Cutie does a good job of segmentation in the prefix, as shown by the highlighted person. During the interjection period, however, the model latches onto the duck in the video, demonstrating false positive predictions and the tarnishing of memory. In the suffix, the model includes parts of the other person's fist in the mask prediction, showing working memory's imperfect re-identification capabilities. Cutie+, on the other hand, does a much better job at segmenting the object. No false positive predictions are made during the interjection period, and the re-identification is much cleaner without any additional pixels.\\

\noindent The objective performance of these models can be evaluated by finding the J\&F score, representing the average of the Jaccard (ground truth and predicted mask overlap) and F1 score. Because each sample contains multiple frame regions, different evaluations are necessary to fully determine the accuracy of the model. The possible evaluation regions are as follows:

\begin{enumerate}
    \item J\&F score over entire video including interjected frames (Table \ref{table:full}).
    \item J\&F score of the 12-frame prefix. This is identical to performance on clean video and is therefore ignored.
    \item Percentage of false-positive predictions during interjection frames. This is simply the percentage of pixels that are predicted to be part of the object. J\&F score cannot be used because the ground truth is all negative, thus there can be no true positive predictions (Table \ref{table:interjection}).
    \item J\&F score of the suffix. This provides an understanding of how good the model is at reidentifying the object and how much the memory has been warped by the interjection frames (Table \ref{table:suffix}).
\end{enumerate}

\begin{table}[h!]
\centering
\begin{tabular}{|c|c|c|c|c|c|}
\hline
\multicolumn{6}{|c|}{Full Video Performance (J\&F Scores)} \\
\hline
Model & 0-Frames (Clean Video) & 4-Frames & 16-Frames & 128-Frames & 512-Frames \\
\hline \hline
XMem & 81.6 & 78.6 & 68.5 & 54.3 & 62.2\\
\hline
XMem+ & 81.6 & \textbf{88.3} & \textbf{93.1} & \textbf{97.8} & \textbf{99.4}\\
\hline \hline
Cutie & 88.1 & 84.9 & 79.5 & 68.3 & 72.1\\
\hline
Cutie+ & 88.1 & \textbf{91.5} & \textbf{95.3} & \textbf{98.5} & \textbf{99.6}\\
\hline \hline
SAM 2 & 88.6 & \textbf{89.0} & 88.8 & 87.9 & 90.1\\
\hline
SAM 2+ & 88.6 & 88.8 & \textbf{95.2} & \textbf{98.5} & \textbf{99.6}\\
\hline
\end{tabular}\\
\caption{Performance of models with and without interjection classifier algorithm on full videos}
\label{table:full}
\end{table}

\begin{table}[h!]
\centering
\begin{tabular}{|c|c|c|c|c|c|}
\hline
\multicolumn{6}{|c|}{Video Interjection Performance (False Positive \%)} \\
\hline
Model & 0-Frames (Clean Video) & 4-Frames & 16-Frames & 128-Frames & 512-Frames \\
\hline \hline
XMem & - & 2.62 & 3.12 & 2.91 & 1.81\\
\hline
XMem+ & - & \textbf{0.00} & \textbf{0.00} & \textbf{0.00} & \textbf{0.00}\\
\hline \hline
Cutie & - & 1.26 & 1.75 & 1.52 & 1.31\\
\hline
Cutie+ & - & \textbf{0.00} & \textbf{0.00} & \textbf{0.00} & \textbf{0.00}\\
\hline \hline
SAM 2 & - & 0.78 & 0.83 & 0.54 & 0.59\\
\hline
SAM 2+ & - & \textbf{0.00} & \textbf{0.00} & \textbf{0.00} & \textbf{0.00}\\
\hline
\end{tabular}\\
\caption{Model false positive rate on interjected frames}
\label{table:interjection}
\end{table}

\begin{table}[h!]
\centering
\begin{tabular}{|c|c|c|c|c|c|}
\hline
\multicolumn{6}{|c|}{Video Suffix Performance (J\&F Scores)} \\
\hline
Model & 0-Frames (Clean Video) & 4-Frames & 16-Frames & 128-Frames & 512-Frames \\
\hline \hline
XMem & - & \textbf{86.3} & 86.1 & 76.9 & 81.2\\
\hline
XMem+ & - & 84.2 & \textbf{86.4} & \textbf{83.8} & \textbf{85.7}\\
\hline \hline
Cutie & - & \textbf{88.9} & 91.0 & 84.9 & 87.0\\
\hline
Cutie+ & - & 88.8 & \textbf{91.4} & \textbf{89.0} & \textbf{90.2}\\
\hline \hline
SAM 2 & - & \textbf{89.9} & 89.6 & 86.1 & 91.1\\
\hline
SAM 2+ & - & 86.5 & \textbf{91.5} & \textbf{90.0} & \textbf{91.2}\\
\hline
\end{tabular}\\
\caption{Performance of models on video suffixes}
\label{table:suffix}
\end{table}

\begin{figure}[h!]
\centering
\includegraphics[width=0.7\linewidth]{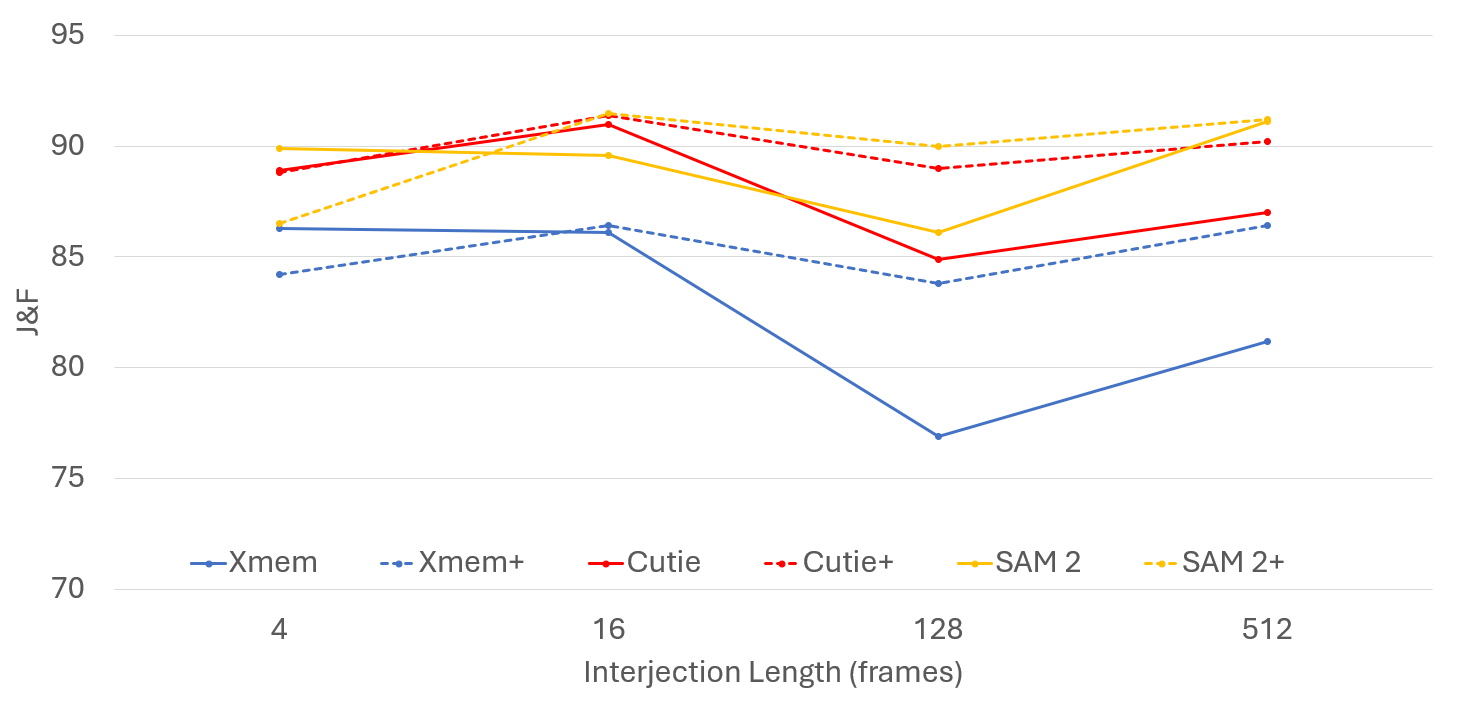}
\caption{\label{fig:suffix} Suffix Performance of Various Models}
\end{figure}

\noindent Figure \ref{fig:suffix} shows a plotted representation of the different models' suffix performance. In nearly all cases, the models with the algorithmic interjection classifier outperforms the original model, with the improvement being increasingly visible with longer interjection periods. This confirms the hypothesis that writing interjection frames into working memory causes quality deterioration over time, thus damaging re-identification performance. By skipping these frames, the working memory can be conserved, and overall video understanding can be maintained over indefinitely long interjection periods.

\pagebreak
\section{Conclusion}

A simple algorithmic change is proposed that can be applied to existing VOS models to improve performance on sudden camera cuts, frame interjections, and extreme context changes. Working memory is inherently susceptible to deterioration over time in the presence of irrelevant frames due to resulting false positive predictions and the writing of these frames into memory. By identifying these irrelevant frames before making predictions, their image embeddings can be skipped, and the working memory can be conserved. Through a series of proposed thresholding functions, very high classification accuracy removes the potential trade off of poor performance on clean video. Future work aims to continue developing patches for working memory issues in video segmentation.

\subsection*{Acknowledgements}

This work was performed under the auspices of the U.S. Department of Energy by Lawrence Livermore National Laboratory under Contract DE-AC52-07NA27344 with funding from the lab and the U.S. Navy. Release number: LLNL-TR-870665

\pagebreak
\bibliographystyle{plain}
\bibliography{bibliography}

\end{document}